\documentclass{article}

\PassOptionsToPackage{square,sort,comma,numbers}{natbib}
\usepackage[preprint]{neurips_2021}

\usepackage[utf8]{inputenc} % allow utf-8 input
\usepackage[T1]{fontenc}    % use 8-bit T1 fonts
\usepackage{hyperref}       % hyperlinks
\usepackage{url}            % simple URL typesetting
\usepackage{booktabs}       % professional-quality tables
\usepackage{amsfonts}       % blackboard math symbols
\usepackage{nicefrac}       % compact symbols for 1/2, etc.
\usepackage{microtype}      % microtypography
\usepackage{xcolor}         % colors
\usepackage{amsmath}
\usepackage{float}
\usepackage{graphicx}
 \usepackage{subfigure} 
\usepackage{multirow}
\usepackage{pythonhighlight}

\title{Analyze and Design Network Architectures by Recursion Formulas}

\author{
  Yilin Liao, Hao Wang, Zhaoran Liu, Haozhe Li and Xinggao Liu\\
  State Key Laboratory of Industry Control Technology\\
  College of Control Science and Engineering\\
  Zhejiang University\\
  Hangzhou 310027, P.R. China \\
  \texttt{ylliao, hwang, zrliu, hzli, lxg @zju.edu.cn} \\
}

\begin{document}

\maketitle

\begin{abstract}
 
  The effectiveness of shortcut/skip-connection has been widely verified, which inspires massive explorations on neural architecture design.
  This work attempts to find an effective way to design new network architectures.
 
  it is discovered that the main difference between network architectures can be reflected in their recursion formulas.
  % COMMENTS: reflect via recursion-> represented via their recursion
  % COMMENTS: It is discovered that the main difference between network architectures can be represented using their recursion formulas.
  Based on this, a methodology is proposed to design novel network architectures from the perspective of mathematical formulas. 

  Afterwards, a case study is provided to generate an improved architecture based on ResNet.
  
  Furthermore, the new architecture is compared with ResNet and then tested on ResNet-based networks. Massive experiments are conducted on CIFAR and ImageNet, which witnesses the significant performance improvements provided by the architecture.

\end{abstract}

\section{Introduction}

Years of practice has proved that convolutional neural network(CNN) is a significantly 
important deep learning method in the field of image recognition.
Many new architectures of CNN have been proposed and have achieved
good performance\cite{mv1, mv2,shuf1,shuf2,inc4,inc1,inc2}.
With the proposal of highway network\cite{highway} and residual network\cite{res}, 
skip-connection and shortcut are widely 
used in deep CNNs\cite{se,dense,resnext,res2net,effi}, 
which results in CNNs with amazing depth.
Residual networks have then achieved subversive performance in a variety of visual 
tasks\cite{res, res2}. 
As a consequence, shortcut has now become an essential component of most deep networks.

The successful application of skip-connection and shortcut has led to a series 
of research explorations on the CNN architecture, e.g., DenseNet\cite{dense}, 
Feedback Network\cite{feedb}, EfficientNet\cite{effi}, etc.
Models designed to improve block performance have also appeared one after another, 
e.g., SEResNet\cite{se}, ResNeXt\cite{resnext}, 
Res2Net\cite{res2net}, etc.
With the emergence of these amazing research work and the reviews\cite{ensem, rev, rev2} 
of previous work, we begin to think about the question that 
\begin{center}
    \emph{Can we find an effective method to design architectures of networks?}\\
    \emph{Can the new network architecture is designed according to our goal?}
\end{center}

In this paper, we first analyze and compare the CNNs with and without shortcut 
--- take the partial derivative of the output of the last layer/block with respect to 
the output of a previous layer/block to make the propagation paths
in the networks be presented in a mathematical formula. 
By comparison, it is discovered that it is the difference 
in network recursion formulas that causes the huge difference between the 
the CNNs without and with shortcut. 
In addition, a new interpretation of the residual network is obtained by partial derivative analysis.

\begin{figure}
    \centering
    \includegraphics[]{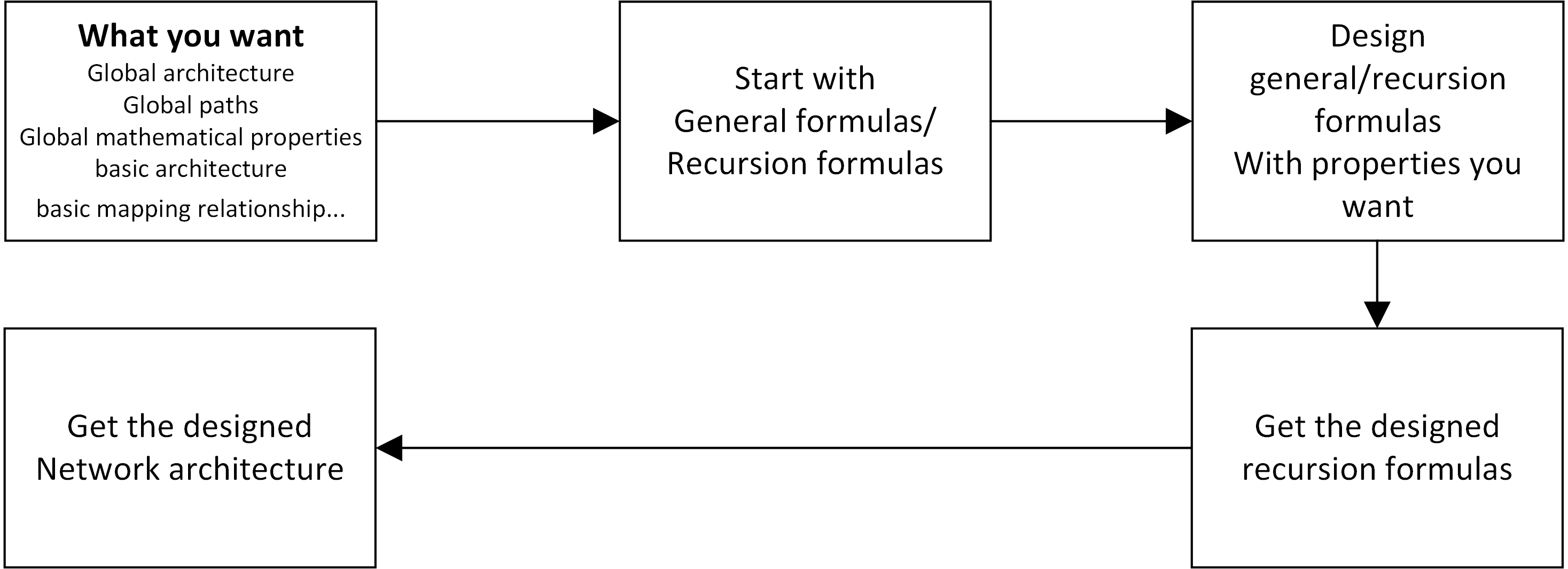}
    \caption{The flow chart of the proposed method}
    \label{fig:flow}
\end{figure}

By the comparative analysis of the two types of networks, 
the significance of recursion formulas to network architecture highlights. 
So, by the reasonable design of network recursion formulas, 
networks with different architectures can be further designed as Figure \ref{fig:flow}.

By the guide of this idea, we take ResNet as an example, 
make some modifications to the general formula of ResNet and  
a new recursion formula is obtained, based on which we design a novel architecture.
We evaluate the new architecture on CIFAR\cite{cifar} and ImageNet\cite{imagenet} data sets,
and it has better performance compared with ResNet.
Further, we test the generality and compatibility of 
the new architecture with other popular networks . 
Experiments show that networks based on the newly designed architecture
 have improvement in performance compared to themselves before.

\section{Related Work}
\paragraph{Highway network}
Highway networks\cite{highway} present shortcut connections with gating functions\cite{lstm}.
The output of each layer of a highway network is defined as\cite{ensem}
\begin{equation}
\label{eqt1}
    y_{i+1} \equiv f_{i+1}\left(y_{i}\right) \cdot t_{i+1}\left(y_{i}\right)+y_{i} \cdot\left(1-t_{i+1}\left(y_{i}\right)\right),
\end{equation}

where $y_{i+1}$ and $y_{i}$ denote the output and input of $i$-th layer,
$t_{i+1}$ denotes the gate function of $(i+1)$-th layer,  $f_{i+1}\left(x\right)$ denotes 
some sequence of convolutions. Eq.\eqref{eqt1} is actually the recursion formula of highway networks.
And highway network is one of the networks to embody the network 
architecture in the network recursion formula.

\paragraph{Residual network}
Identity mapping allows neural networks to reach a depth that was unimaginable before.
In the analysis of the residual network\cite{res,res2}, the recursion formula plays a vital role.
Residual unit performs the following computation:
\begin{equation}
    \begin{array}{c}\mathbf{y}_{l}=h\left(\mathrm{x}_{l}\right)+F\left(\mathrm{x}_{l}, \omega_{l}\right) \\ \mathrm{x}_{l+1}=fp\left(\mathrm{y}_{l}\right)\end{array},
\end{equation}
where, $\mathrm{x}_{l}$ and  $\mathrm{x}_{l+1}$ are the input and output the $l$-th residual unit, 
$\omega_{l}$ is a set of weights (and biases) associated with the $l$-th residual unit.
The function $fp$ is the operation after element-wise addition. 
The function $h$ is set as an identity mapping:$h\left(\mathrm{x}_{l}\right)=\mathrm{x}_{l}$.

Furthermore, partial derivation is used to 
analyze the relationship between the outputs of units.
Paper\cite{res2} analyzes the meaning of each item 
in the partial derivative formula in the network, 
and it is proved in the meanwhile that
shortcut ensures that information can be directly 
propagated back to any shallower unit. 
Our work is parallel to this and gives a further view 
on this basis that the recursion formula is the embodiment of architecture
and partial derivative formulas can be 
used to study the propagation paths in the network.

\paragraph{Study on the architecture of ResNet}
Paper\cite{ensem} analyzes the architecture of residual networks and 
comes to a conclusion that residual networks behave like 
ensembles of relatively shallow networks. This work shows 
the propagation paths that residual network contains by 
forward derivation of the recursion formula. 
We come to a similar conclusion by the partial derivative of 
the output of the last block with respect to 
the output of a previous block.

\section{Analysis}
\label{ana}
\subsection{Analysis of Traditional CNNs}
\label{res}
For the traditional convolutional neural networks\cite{cnn}, the recursion formula is:
\begin{equation}
    X_{i}=g_{i}\left[F\left(X_{i-1}, \theta_{i}\right)\right], 0<i<L+1
\end{equation}
 where $L$ is the total number of layers,
 $X_{i}$ denotes the output of $i$-th layer, $g_{i}$ denotes the activation function of $i$-th layer ,  
 $F$ denotes the mapping of the layer from input to output and 
$\theta_{i}$ denotes the parameter of $i$-th layer.

In order to make the propagation paths in the network easy to understand,
we take the partial derivative of the output of the last block with respect to 
the output of a previous block\cite{res2}:
\begin{align}
&\frac{\partial X_{L}}{\partial X_{L-1}}=g_{L}^{\prime} \cdot W_{L} \label{c1}\\
&\frac{\partial X_{L}}{\partial X_{L-2}}=\frac{\partial X_{L}}{\partial X_{L-1}} \frac{\partial X_{L-1}}{\partial X_{L-2}}=g_{L}^{\prime} \cdot W_{L} \cdot g_{L-1}^{\prime} \cdot W_{L-1} \label{c2} \\
&... \notag \\
&\frac{\partial X_{L}}{\partial X_{L-i}}=\frac{\partial X_{L}}{\partial X_{L-1}} \frac{\partial X_{L-1}}{\partial X_{L-2}} \ldots \frac{\partial X_{L-i+1}}{\partial X_{L-i}}=g_{L}^{\prime} W_{L} \cdot g_{L-1}^{\prime} W_{L-1} \cdot \ldots \cdot g_{L-i+1}^{\prime} W_{L-i+1} \label{c3} \\
&where\quad0<i<L+1. \notag
\end{align}

where $W_{i} =\frac{\partial F\left(X_{i-1}, \theta_{i}\right)}{\partial X_{i-1}}$,
$W_{i}$ represents the mapping from input to ouput of the $i$-th layer and 
$g^{\prime}$ denotes the derivative of activation function $g$.

There is a idea : \\
If the partial derivative of the output of the last block(or layer) 
with respect to the output of $block_{j}/or \ layer_{j}$ contains $W_{k} (j<k<L+1)$, 
information could go through $block_{k}/or \ layer_{k}$ in the course of 
the propagation from $block_{j}/or \ layer_{j}$ to the last block(or layer).

\begin{figure}
  \centering
  \subfigure[old CNNs]{
    \label{fig:subfig:a} %% label for first subfigure
    \includegraphics[width=0.75in]{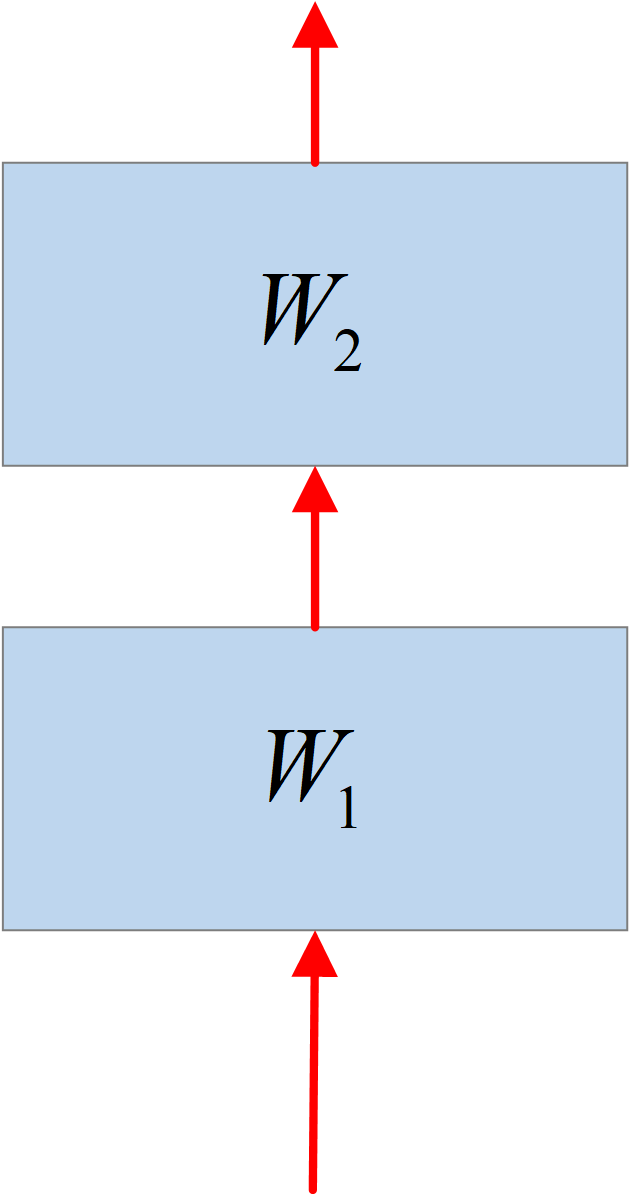}}
  \qquad
  \subfigure[residual networks $\left(1+W_{1}\right)\left(1+W_{2}\right)=1+W_{1}+W_{2}+W_{1} W_{2}$]{
    \label{fig:subfig:b} %% label for second subfigure
    \includegraphics[width=4.0in]{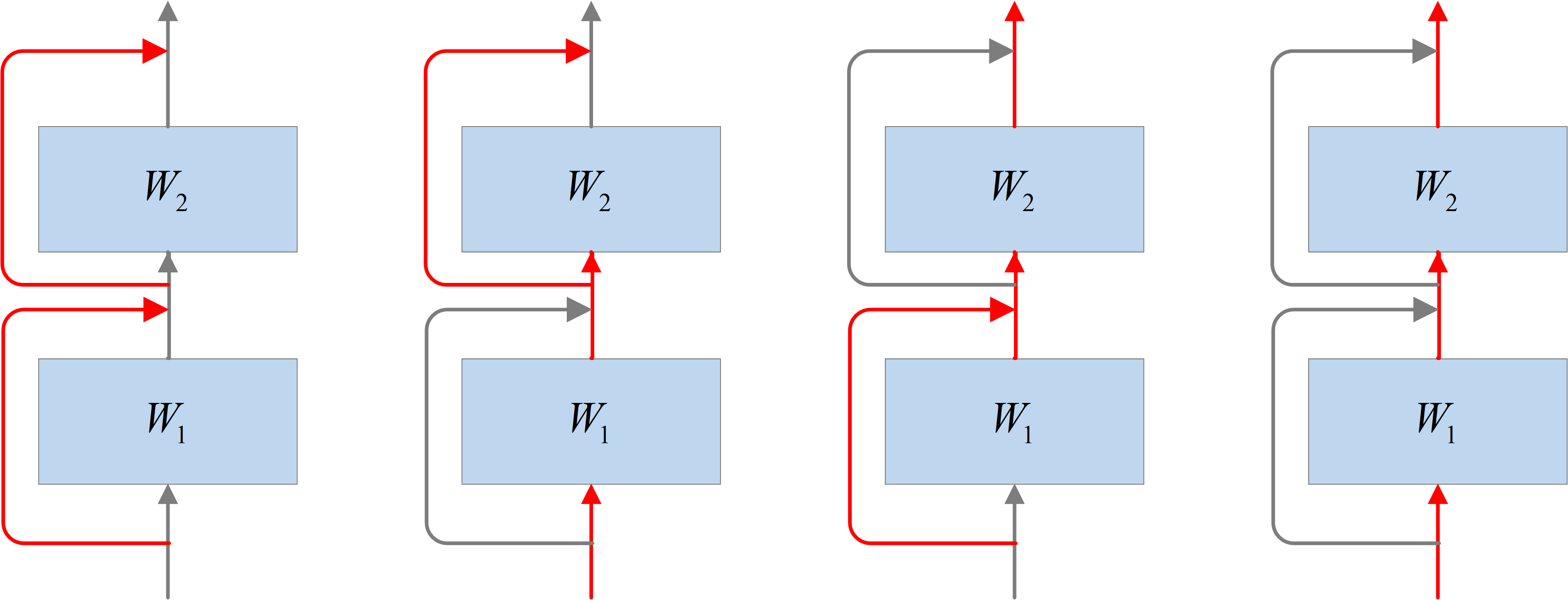}}
  \caption{Propagation paths in networks(red lines represent optional paths)}
  \label{fig:subfig} 
\end{figure}

With this idea, review Eq.\eqref{c3}: the network is stacked by
$W_{i}$, which means the network is built up layer by layer.
Also, we can clearly see that there is only one path from 
the output of $layer_{L-i}$ to the output of $layer_{L}$ 
--- go through $layer_{L-i+1}$, $layer_{L-i+2}$, ..., 
till $layer_{L}$ as Figure \ref{fig:subfig:a}. 

\subsection{Analysis of Residual Networks}
\label{res1}
For the residual networks, the recursion formula is\cite{res,res2}:
\begin{equation}
    X_{i}=g_{i}\left[X_{i-1}+F\left(X_{i-1}, \theta_{i}\right)\right], 0<i<L+1
\end{equation}
where $L$ is the total number of blocks.

Similarly, in order to make the propagation paths in the network easy to understand\cite{res2},
we take the partial derivative of the output of the last block with respect to 
the output of a previous block and get the following fomula:
\begin{align} 
\frac{\partial X_{L}}{\partial X_{L-i}} &=\frac{\partial X_{L}}{\partial X_{L-1}} \cdot \frac{\partial X_{L-1}}{\partial X_{L-2}} \cdot \ldots \cdot \frac{\partial X_{L-i+1}}{\partial X_{L-i}} \\ &=g_{L}^{\prime}\left(1+W_{L}\right) \cdot g_{L-1}^{\prime}\left(1+W_{L-1}\right) \cdot \ldots \cdot g_{L-i+1}^{\prime}\left(1+W_{L-i+1}\right) \label{3}
\end{align}

\paragraph{The new view of residual networks}
To make a better comprehension of the architecture of the network,
ignore the activation function in the Eq.\eqref{3} 
(activation function does not affect the architecture of a network):
\begin{align}
\frac{\partial X_{L}}{\partial X_{L-i}} &=\left(1+W_{L}\right)\left(1+W_{L-1}\right) \cdot \ldots \cdot\left(1+W_{L-i+1}\right) \label{t7}\\
&=1+\mathrm{S}_{i}^{1} W+\mathrm{S}_{i}^{2} W+\ldots+\mathrm{S}_{i}^{i-2} W+
\mathrm{S}_{i}^{i-1} W+\mathrm{S}_{i}^{i} W \label{t8}
\end{align}
$\mathrm{S}_{i}^{k}W$ represents all combinations that any k terms selected from $W$ to multiply,
where $W=\left(W_{L}, W_{L-1}, W_{L-2}, \ldots, W_{L-i+1}\right)$.
For example, $\mathrm{S}_{3}^{2}({W}_{L}, {W}_{L-1}, {W}_{L-2}) = {W}_{L}\cdot{W}_{L-1} + 
{W}_{L}\cdot{W}_{L-2} + {W}_{L-1}\cdot{W}_{L-2}$.

Similarly to section \ref{res}, according to Eq.\eqref{t7}, we know that ResNets
are stacked by $1 + W_{i}$, which means the basic unit of ResNets
consists of a shortcut and some sequence of convolutions.

In Eq.\eqref{t8}, addition means parallel connection and 
multiplication means series connection. And, 
each multiplication term represents a path.
For example, in $({W}_{L}\cdot{W}_{L-1} + {W}_{L}\cdot{W}_{L-2} + {W}_{L-1}\cdot{W}_{L-2})$,
there are three paths that go through two blocks, 
which are ${W}_{L}\cdot{W}_{L-1}$, ${W}_{L}\cdot{W}_{L-2}$ and ${W}_{L-1}\cdot{W}_{L-2}$.

So, according to Eq.\eqref{t8} and the definition of $\mathrm{S}_{i}^{k}W$,
we can clearly see the propagation paths 
from the output of $block_{L-i}$ to the output of $block_{L}$ as Figure \ref{fig:subfig:b}:\\
There is one path from ${block}_{L-i}$ straight to ${block}_{L}$ \\
There are $\mathrm{C}_{1}^{i}$ paths from ${block}_{L-i}$ to ${block}_{L}$ through 1 block \\
There are $\mathrm{C}_{2}^{i}$ paths from ${block}_{L-i}$ to ${block}_{L}$ through 2 blocks \\
... \\
There are $\mathrm{C}_{i-1}^{i}$ paths from ${block}_{L-i}$ to ${block}_{L}$ through $i-1$ blocks \\
There is $\mathrm{C}_{i}^{i}$ path from ${block}_{L-i}$ to ${block}_{L}$ through $i$ blocks\\
where $C_{k}^{i}=\frac{i !}{k !(i-k) !}$, and $!$ represents factorial.

\section{Method}
According to section \ref{ana}, by the comparison
, the biggest difference between residual networks and the previous networks
is the recursion formula --- It is the small change of "+ 1" in the recursion formula
that makes the entire network more powerful by recursion and multiplication.

Inspired by this, the architecture design of a network can 
start with the recursion formulas. 
Thus, we can study the properties of the network from the math nature, 
and explore better network architectures and more possibilities.
Architectures designed in this way will be more interpretable in math.

The ultimate goal is the expected recursion formulas.
There may be various ways to get them.
Two ways to design new architectures are provided here:
\paragraph{Start with general formulas} 
By the analysis of section \ref{res} and \ref{res1}, it can be found that
general formulas reflect the propagation paths in the network.
Thus, if we want to change or design propagation paths 
from the perspective of the  whole network, we can start with 
the general formulas like Eq.\eqref{c3}, Eq.\eqref{3} or 
$X_{L}=F\left(X_{L}, X_{L-1}, X_{L-2}, \ldots,X_{0}\right)$, etc
, $F$ is the expected function.
According to the designed general formulas, the recursion formulas can be obtained.
Then, the new architecture can be designed. The flow chart of this method is 
shown as Figure \ref{fig:flow}.

\paragraph{Start with recursion formulas}
Recursion formulas represents the basic units in the network, which 
reflect the basic architecture of the network and 
the mapping relationship between the input and output of the basic units.
Thus, if we want a certain basic architecture or mapping relationship 
like $X_{i}=F\left(X_{i}, X_{i-1}, X_{i-2}, \ldots\right)$ or  
$X_{i}=F\left(W_{i}\right) \cdot X_{i-1}$, etc,  
we can directly start with recursion formulas, 
based on which the new architecture can be designed.
The flow chart of this method is shown as Figure \ref{fig:flow}.

\section{Example}
\label{example}
\subsection{Get the recursion formulas}
\label{case_prof}
According to section \ref{res}, consider the following problems\cite{ensem} in ResNet:
ResNet tends to take the shorter paths during training.
Pathways of the same length have different widths and different expressive power.
Too many optional paths will cause redundancy.

In terms of the above problems, we make some modifications to the Eq.\eqref{t7}:
\begin{align}
&\frac{\partial X_{L}}{\partial X_{L-i}}=1+\mathrm{W}_{L}+\mathrm{W}_{L} \cdot \mathrm{W}_{L-1}+\ldots+\mathrm{W}_{L} \cdot \mathrm{W}_{L-1} \cdot ... \cdot \mathrm{W}_{L-i+1} \label{eq4} \\
&where\quad0<i<L. \notag
\end{align}
Eq.\eqref{eq4} guarantees that 
there is only one path per length from $X_{L-i}$ to $X_{L}$, which alleviates the redundancy.
Besides, paths go through the the widest blocks ${W}_{L} \cdot {W}_{L-1} \cdot \ldots \cdot {W}_{L-i+1}$(for ${W}_{i}$, the larger $i$, the wider ${block}_{i}$),  
which ensures that the paths have the strongest expressive power.
From ${W}_{L} \cdot {W}_{L-1} \cdot \ldots \cdot {W}_{L-i+1}$, 
it can also be found that longer paths are based on shorter paths.

For convenience, let:
\begin{equation}
    \frac{\partial X_{L}}{\partial X_{L-i}}=f_{L-i}
\end{equation}
Bring $\frac{\partial X_{L}}{\partial X_{L-i}}=f_{L-i}$ into Eq.\eqref{eq4},
then we can get the recursion formula from them:
\begin{align} 
f_{L-i} &=f_{L-i+1}+W_{L-i+1} \cdot\left(f_{L-i+1}-f_{L-i+2}\right) \notag\\ 
&=\left(1+W_{L-i+1}\right) \cdot f_{L-i+1}-W_{L-i+1} \cdot f_{L-i+2} \label{rec}\\ 
&where\quad1<i<L+1. \notag
\end{align} 

In Eq.\eqref{rec}, it can be seen that the output of the $(L-i)$-th block is not only related to the $(L-i+1)$-th block, 
but also related to the $(L-i+2)$-th block.
This gives the basic block a memory capacity.

Then, bring $f_{L-i} = \frac{\partial X_{L}}{\partial X_{L-i}}$ into Eq.\eqref{rec}:
\begin{align}
&\frac{\partial X_{L}}{\partial X_{L-i}}=\left(1+W_{L-i+1}\right) \cdot \frac{\partial X_{L}}{\partial X_{L-i+1}}-W_{L-i+1} \cdot \frac{\partial X_{L}}{\partial X_{L-i+2}} \label{case_rec}\\
&where \quad 1<i<L+1  \notag
\end{align}

An important method to further derive the partial derivative formula is the chain rule.
Eq.\eqref{case_rec} is supposed to be in the form of chain rule like Eq.\eqref{eq_chain}, 
so that it makes sense: 
\begin{align}
\frac{\partial X_{L}}{\partial X_{L-i}} &=\frac{\partial X_{L}}{\partial X_{L-i+2}} \frac{\partial X_{L-i+2}}{\partial X_{L-i}} \label{eq_chain}
\end{align}

Eq.\eqref{case_rec} can be regarded as the derivation from Eq.\eqref{eq_chain} or 
the expanded form of Eq.\eqref{eq_chain}. Make them equal and we can get the following equtaion:
\begin{equation}
\label{deeq}
\frac{\partial X_{L-i+2}}{\partial X_{L-i}}=\left(1+\mathrm{W}_{L-i+1}\right) \frac{\partial X_{L-i+2}}{\partial X_{L-i+1}}-\mathrm{W}_{L-i+1} 
\end{equation}

According to Eq.\eqref{deeq}, the inferences that
$\frac{\partial X_{L-i+1}}{\partial X_{L-i}}=1+\mathrm{W}_{L-i+1}$ 
and $X_{L-i+2}$ consists of the term of 
($-{W}_{L-i+1} \cdot X_{L-i}$) can be further obtained. Based on the inferences, we can get the following recursion formulas, which are good demonstration of the relationship between the blocks in the network and make it easy to learn about the overall architecture of the network:
\begin{align}
&X_{L}=\left(1+\mathrm{W}_{L}\right) \cdot X_{L-1}-\mathrm{W}_{L-1} \cdot X_{L-2}  \label{case_frec0}\\
&X_{L-1}=\left(1+\mathrm{W}_{L-1}\right) \cdot X_{L-2}-\mathrm{W}_{L-2} \cdot X_{L-3}  \\
&... \notag \\
&X_{L-i}=\left(1+\mathrm{W}_{L-i}\right) \cdot X_{L-i-1}-\mathrm{W}_{L-i-1} \cdot X_{L-i-2} \label{case_frec} \\
&where\quad 0<i<L-1 \notag \\
&... \notag \\
&X_{1}=\left(1+\mathrm{W}_{1}\right) \cdot X_{0} \notag
\end{align}

\subsection{Design the architecture}
\begin{figure}
    \centering
    \includegraphics[scale=2]{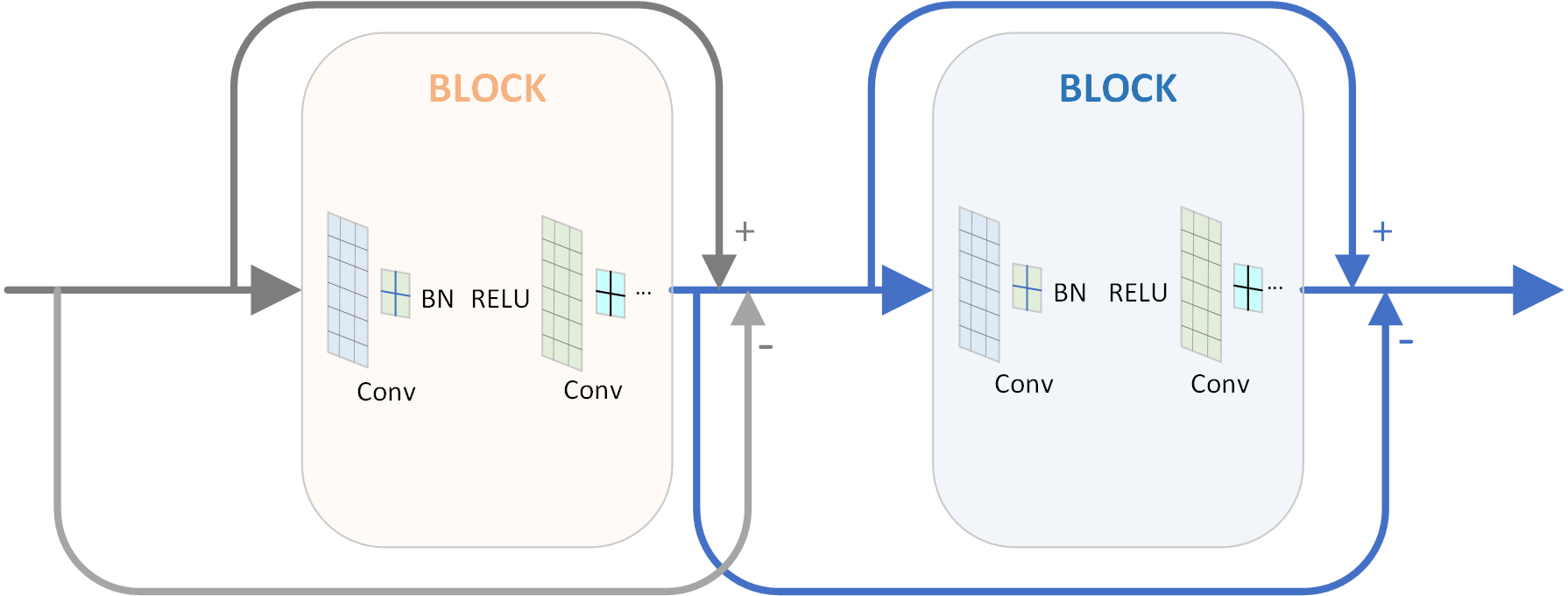}
    \caption{The newly designed architecture}
    \label{fig:pic1}
\end{figure}

According to the Eq.\eqref{case_frec}, 
it is easy to find that the equation has two parts:$\left(1+W_{L-i+1}\right) \cdot f_{L-i+1}$ 
which is the same as ResNet and $W_{L-i+1} \cdot f_{L-i+2}$ which 
is the output of the convolutional sequences in the previous block.
We can design the architecture of the network as Figure \ref{fig:pic1}. 
This example starts with general formulas that we want. Then,
according to the general formulas, we get new recursion formulas, based on which 
we design the new architecture.

Note that the architecture must be designed according to recursion formulas like Eq.\eqref{case_frec}.
Only in this way can the designed architecture has the expected propagation paths.
Considering $X_{1}=\left(1+\mathrm{W}_{1}\right) X_{0}$ and the Eq.\eqref{case_frec}, 
if $X_{1}=\left(1+\mathrm{W}_{1}\right) X_{0}$ is brought into the recursion formulas,
Eq.\eqref{case_frec0}-\eqref{case_frec} will change into the following form:
\begin{align}
&X_{q}=W_{q} X_{q-1}+X_{0} \label{ll1}\\
&where \quad 1<q<L+1 \notag
\end{align}
Although, when $X_{1}=\left(1+\mathrm{W}_{1}\right) X_{0}$ is given, 
Eq.\eqref{case_frec0}-\eqref{case_frec} and Eq.\eqref{ll1} are equivalent.
The architectures designed according to them are quite different.
And it can be found that if the architecture is designed according to Eq.\eqref{ll1},  
the output of the previous block cannot directly propagate to the next block 
and only the input of the network can propagate across layers, 
which is not in line with our idea.
In addition, architectures designed by Eq.\eqref{case_frec} and Eq.\eqref{ll1}
have quite different performance in experiments. The latter perform much worse
--- slower convergence and worse accuracy.

The propagation paths should be emphasized in the design process.
So, the network architecture is supposed to be designed according to the original recursion formulas 
, rather than the recursion formulas after the initial values are brought in.

\subsection{Experiment}
\label{experiment}
In this section, names plus $s$ of the networks denote corresponding networks with the new architecture.
GPUs used are four TITAN RTXs. The complete experimental data obtained by using 
different random seeds is given in the appendix \ref{ap2}.
\subsubsection{Comparison with ResNet}
\label{comres}
The performances of ResNet18/34/50/101\cite{res} and 
the corresponding networks with the new architecture are compared on the test sets.
Because the new architecture is designed on the basis of ResNet, 
the hyperparameters of the corresponding networks are completely consistent.

On CIFAR data sets, all the networks are trained with  
momentum stochastic gradient descent\cite{sgdm} optimizer with a learning rate of 0.1, 
momentum of 0.9, and the weight decay $5\times10^{-4}$. 
Besides, the scheduler used is cosine annealing\cite{cos} 
, the number of training epoch is 200 and batch size is 128.

On ImageNet data set, all the networks are trained with  
momentum stochastic gradient descent\cite{sgdm} optimizer with a learning rate of 0.1, 
momentum of 0.9, and weight decay $5\times10^{-4}$. 
Besides, the scheduler used is MultiStepLR, the learning rate decays by a factor of 10 every 30 epochs, 
the number of training epoch is 90 and batch size is 256.

Table \ref{table1} shows the results of ResNets and the corresponding 
networks with newly designed architecture in our experiments.
The networks with newly designed architecture achieve better performance than ResNets.
And there is a steady improvement in accuracy on different baselines.
Restricting information to the paths with strongest expressive power 
does improve network performance.

\begin{table}
  \caption{The results of the newly designed network on test sets}
  \label{table1}
  \centering
  \begin{tabular}{ccccccc}
    \toprule
    \multirow{2}{*}{Method} & \multicolumn{2}{c}{CIFAR10} & \multicolumn{2}{c}{CIFAR100} 
    & \multicolumn{2}{c}{ImageNet}\\
    \cmidrule(r){2-3} \cmidrule(r){4-5} \cmidrule(r){6-7}
    & top-1(\%) & top-5(\%)
    & top-1(\%) & top-5(\%)
    & top-1(\%) & top-5(\%) \\
    \midrule
    ResNet18   &94.32          &99.76     &76.60          &93.36      &68.63          &88.58\\
    ResNet18s  &\textbf{94.70} &99.80     &76.57          &93.41      &68.51          &88.41 \\
    ResNet34   &94.75          &99.78     &77.09          &93.27      &72.14          &90.42 \\
    ResNet34s  &\textbf{95.10} &99.88     &\textbf{77.48} &93.77      &\textbf{72.38}   &90.77\\
    ResNet50   &95.00          &99.88    &78.37          &94.26       &75.13          &92.36 \\
    ResNet50s  &\textbf{95.29} &99.85    &\textbf{78.65} &94.32       &\textbf{75.40}  &92.52 \\
    ResNet101  &94.89          &99.83    &77.67          &93.79       &76.51          &92.96 \\
    ResNet101s &\textbf{95.01} &99.86    &\textbf{78.56} &93.86       &\textbf{76.73}     &93.20 \\
    \bottomrule
  \end{tabular}
\end{table}

\subsubsection{Friedman Test \& Nemenyi Test}
\label{fn}
In order to verify the significance of the new architecture\cite{comp},
we conduct Friedman test and Nemenyi test on the networks\cite{comp, fredm, ntest}.
The specific principles of this method are in the appendix \ref{ap1}.

\begin{figure}
    \centering
    \includegraphics{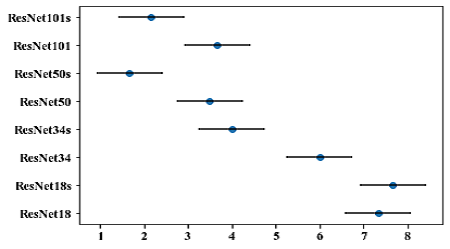}
    \caption{Friedman graph}
    \label{fig:picf}
\end{figure}

According to the experimental data on CIFAR10, CIFAR100 and ImageNet, Friedman graph can be drawn as Figure \ref{fig:picf}, where the horizontal axis value represents the ranking.
The horizontal line in the graph denotes the ranking of the accuracy of the corresponding model in vertical axis. The solid origin denotes the mean value of the ranking on different data sets.

In the Friedman graph, if the two methods have no overlapping area, 
it proves that the performance of two methods is significantly different\cite{comp}.
It can be seen that ResNet34/50/101 and ResNet34s/50s/101s
have no overlapping area respectively. ResNet34s/50s/101s 
have better performance than  ResNet34/50/101 on both CIFAR and ImageNet data sets.
Also, it is easy to find that 
the performance of ResNet34s is comparable to that of ResNet50 and is better than
that of ResNet101 on CIFAR data sets. 
The performance of ResNet50s far exceeds others on CIFAR data sets.
It proves that the performance of the new architecture 
has been significantly improved compared to ResNet.

\subsubsection{Further experiments}
\label{FE}
To validate the generality and compatibility of the proposed 
architecture with other popular neural networks, 
additional experiments are conducted with the 3 baselines: 
SEResNet\cite{se}, ResNeXt\cite{resnext}, Res2NeXt\cite{res2net}.
In table \ref{table2}, ResNeXt2 and ResNeXt4 denote $ResNeXt29(2\times64d)$ and $ResNeXt29(4\times64d)$
respectively\cite{resnext}.
Res2NeXt2 and Res2NeXt4 denote $Res2NeXt29(6c\times24w\times4scales)$ and 
$Res2NeXt29(8c\times25w\times4scales)$ respectively\cite{res2net}.

\begin{table}
  \caption{Results of other networks with the newly designed architecture}
  \label{table2}
  \centering
  \begin{tabular}{ccccccc}
    \toprule
    \multirow{2}{*}{Method} & \multicolumn{1}{c}{CIFAR10} & \multicolumn{1}{c}{CIFAR100} 
    &\multirow{2}{*}{Method} & \multicolumn{1}{c}{CIFAR10} & \multicolumn{1}{c}{CIFAR100}\\
    \cmidrule(r){2-3} \cmidrule(r){5-6} 
    & top-1(\%)  & top-1(\%)  &  & top-1(\%)  & top-1(\%) \\
    \midrule
    SEResNet18   &95.29                   &78.71      
    &SEResNet18s &\textbf{95.47}          &78.65                     \\
    SEResNet34   &95.58                   &79.05       
    &SEResNet34s &\textbf{95.86}          &\textbf{80.15}            \\
    SEResNet50   &95.57                   &79.30      
    &SEResNet50s  &\textbf{95.85}          &\textbf{80.36}           \\
    SEResNet101  &95.51                   &79.36       
    &SEResNet101s &\textbf{95.80}         &\textbf{80.09}            \\
    ResNeXt2     &95.67                   &79.39       
    &ResNeXt2s   &95.55                   &\textbf{79.49}            \\ 
    ResNeXt4     &96.53                   &80.99    
    &ResNeXt4s  &96.53                    &\textbf{81.06}            \\ 
    Res2NeXt2    &95.75                   &81.14      
    &Res2NeXt2s  &\textbf{96.43}          &\textbf{82.22}            \\ 
    Res2NeXt4    &96.16                  &79.27       
    &Res2NeXt4s  &96.09                   &\textbf{82.49}            \\ 
    \bottomrule
  \end{tabular}
\end{table}

\begin{table}
  \caption{Comparison on ImageNet}
  \label{table3}
  \centering
  \begin{tabular}{ccccc}
    \toprule
    \multirow{2}{*}{Method} & \multicolumn{1}{c}{ImageNet}  
    &\multirow{2}{*}{Method} & \multicolumn{1}{c}{ImageNet} \\
    \cmidrule(r){2-2} \cmidrule(r){4-4} 
    & top-1(\%)  &  & top-1(\%)  \\
    \midrule
    SEResNet34   &68.624                        
    &SEResNet34s &\textbf{69.182}                      \\
    SEResNet50   &71.532                        
    &SEResNet50s  &\textbf{72.436}                   \\
    ResNeXt2     &67.591                        
    &ResNeXt2s   &\textbf{68.412}                            \\ 
    ResNeXt4     &72.390                   
    &ResNeXt4s  &\textbf{73.264}                             \\
    \bottomrule
  \end{tabular}
\end{table}

Table \ref{table2} gives the results of networks that take ResNet and the new architecture as architecture respectively.
The newly designed architecture comprehensively improves 
the performance of SEResNet networks, which shows that 
the new architecture has a good compatibility with the attention mechanism.
Through the comparison between ResNeXt/Res2NeXt and ResNeXts/Res2NeXts,
the new architecture can achieve better training effect 
when the model is complex and difficult to train.
At the same time, the new architecture tends to achieve better performance 
on complex tasks.

From another perspective, this architecture 
increases the average width of paths in each length,
which also proves the importance of network width.
We can come to a conclusion that
appropriate increase in width can improve network performance
\cite{wod, wdlc}.

\subsubsection{Experiment summary}
\label{ex-an}
In this example, additional cross-layer connections 
will increase the number of parameters\cite{res}.
We do not focus on reducing the model size in this example since 
what we want to show is the design process and design ideas.
The model sizes and computational complexity are given in appendix \ref{appm}.
On the one hand, this example can be considered as architecture optimization, 
which picks out the paths with the strongest expressive power and 
increases the average width of each length path;
on the other hand, this example can be considered as a method that
increases the number of network parameters to stably improves the network performance 
by additional cross-layer connections while keeping the network depth and feature map channels unchanged.

% Although experiments show that this method of parameter increase 
% can bring general performance improvements in different networks, 
% it’s better that there is no parameter increase.

\section{Conclusion}
In this paper, we show the effect of recursion formulas on network architectures and
show the propagation paths in networks by taking the partial derivative 
of the output of the last block/layer with respect to the output of a previous block/layer.
In the meanwhile, a mathematical method to design network architectures is proposed.
Based on the method, a new architecture is designed as an example.
The entire design process shows how the theory guides the experiment.
Comparative experiments are conducted on CIFAR and ImageNet, which shows 
that the network with the newly designed architecture
has better performance than ResNet and
the newly designed architecture has the generality and compatibility 
with other popular neural networks.
This verifies the feasibility of the proposed method.

\section{Future Work and Broader Impact}
\label{bi}

\paragraph{Future work} This work is an attempt to design the network architecture completely from the perspective of mathematical formulas. In the future, we will continue to improve the theoretical system of using mathematics to solve neural network problems. We hope that through our efforts, we can make the related research of artificial intelligence more mathematical and theoretical, rather than relying more on genius inspiration and a large number of experimental trial and error. Mathematical theory, physical theory and information theory are all great treasures, and they all have a solid theoretical foundation. The analysis methods in these fields can be used for reference. The development of artificial intelligence should not be independent of them, but should be integrated with them.

\paragraph{Broader impact} We analyze the architectures of networks and propagation paths in networks
by mathematical formulas, which makes the network architecture more 
interpretable and easier to understand.
This method to design architectures of networks makes
the design of the network architecture have a mathematical theoretical basis.
This promotes the combination of network architecture design and mathematics.
This may also promote the emergence of more new network architectures in the future.
The negative outcomes result from the increasing accuracy of image recognition 
and more used surveillance cameras. 
That may raise the risk of privacy breaches and other security issues, 
which may put everyone under monitoring. 

% \section{Acknowledgments and Disclosure of Funding}
% This work is supported by State Key Laboratory of Industry Control Technology,
%   College of Control Science and Engineering,
%   Zhejiang University and their supports are thereby acknowledged.

%\section*{References}

\appendix

\section{Appendix}
\subsection{Friedman test and Nemenyi test}
\label{ap1}
For $k$ algorithms and $N$ data sets, 
first get the test performance results of each algorithm on each data set, 
and then sort them according to the performance results from good to bad, 
and give the order value $1, 2, …, k$. 
If the performance results of multiple algorithms are the same, 
they are equally divided into rank values.
Assuming the average rank value of the $i$-th algorithm 
is ${r}_{i}$, then ${r}_{i}$ obeys the normal distribution:
\begin{align}
&\tau_{x^{2}}=\frac{12 N}{k(k-1)}\left(\sum_{i=1}^{k} r_{i}^{2}-\frac{k(k+1)^{2}}{4}\right) \notag \\ &\tau_{\mathrm{F}}=\frac{(N-1) \tau_{x^{2}}}{N(k-1)-\tau_{x^{2}}} \notag
\end{align}
where, $\tau_{F}$ Obey the $F$ distribution 
with degrees of freedom $k-1$ and $(k-1)(N-1)$.

If the hypothesis of "all algorithms have the same performance" is rejected, 
it means that the performance of the algorithms is significantly different, 
and then a follow-up test is required.

The Nemenyi test calculates the critical value range $CD$ of the average ordinal difference:
\[CD=q_{\alpha} \sqrt{\frac{k(k+1)}{6 N}}\]
The $q_{\alpha}$ values are listed in the table \ref{tableq}.

\begin{table}[H]
  \caption{$q_{\alpha}$ values}
  \label{tableq}
  \centering
  \begin{tabular}{ccccccccccc}
    \toprule
    \multirow{2}{*}{$\alpha$} & \multicolumn{9}{c}{k algorithms}\\
    \cmidrule(r){2-10} 
    &2 &3 &4 &5 &6 &7 &8 &9 &10 \\
    \midrule
    0.05   &1.960  &2.334  &2.569 &2.728 &2.850 &2.949 &3.031 &3.102 &3.164\\
    0.10  &1.645  &2.052  &2.291 &2.459 &2.589 &2.693 &2.780 &2.855 &2.920\\
    \bottomrule
  \end{tabular}
\end{table}

If the rank value difference of any two algorithms is bigger than $CD$, 
there is a significant difference in performance of the two algorithms.

\subsection{Experiment Datas}
\label{ap2}
\begin{figure}[H]
    \centering
    \includegraphics[scale = 0.4]{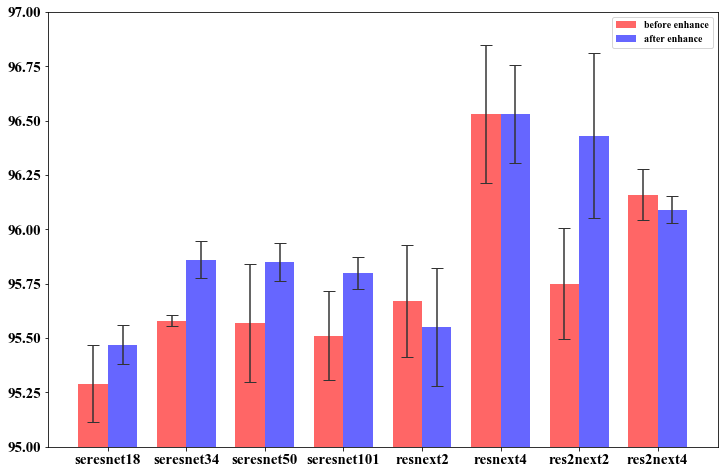}
    \caption{CIFAR10}
    \label{fig:c10}
\end{figure}

\begin{figure}[H]
    \centering
    \includegraphics[scale = 0.4]{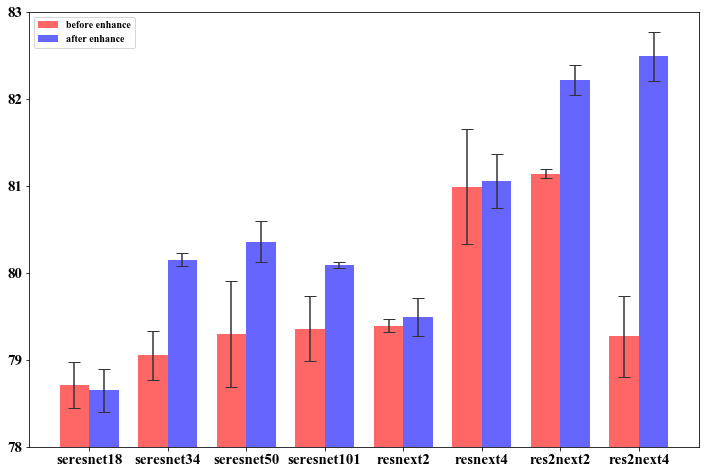}
    \caption{CIFAR100}
    \label{fig:c100}
\end{figure}

The complete experiment data mentioned in section \ref{FE} is
in table \ref{tableap}, table \ref{tableap1} and table \ref{tableap2}.
The histograms with error bars are in Figure \ref{fig:c10} and Figure \ref{fig:c100}.
Complete data on ImageNet is being collected.

\begin{table}[H]
  \caption{Seed = 0}
  \label{tableap}
  \centering
  \begin{tabular}{cccccc}
    \toprule
    \multirow{2}{*}{Method} & \multicolumn{1}{c}{CIFAR10} & \multicolumn{1}{c}{CIFAR100} 
    &\multirow{2}{*}{Method} & \multicolumn{1}{c}{CIFAR10} & \multicolumn{1}{c}{CIFAR100}\\
    \cmidrule(r){2-3} \cmidrule(r){5-6} 
    & top-1(\%)  & top-1(\%)  &  & top-1(\%)  & top-1(\%) \\
    \midrule
    SEResNet18   &95.42           &79.01          
    &SEResNet18s  &95.54          &78.91           \\
    SEResNet34   &95.62           &79.34           
    &SEResNet34s  &95.91          &80.05           \\
    SEResNet50   &95.52           &80.10           
    &SEResNet50s  &95.83          &80.65           \\
    SEResNet101  &95.42           &79.09           
    &SEResNet101s &95.79          &80.09           \\
    ResNeXt2     &95.98          &79.31            
    &ResNeXt2s    &95.90         &79.72            \\ 
    ResNeXt4     &96.36          &81.59            
    &ResNeXt4s    &96.44         &81.45            \\ 
    Res2NeXt2    &95.40          &81.14             
    &Res2NeXt2s   &96.16         &82.43            \\ 
    Res2NeXt4    &96.32          &78.66             
    &Res2NeXt4s   &96.02         &82.09            \\ 
    \bottomrule
  \end{tabular}
\end{table}

\begin{table}[H]
  \caption{Seed = 617}
  \label{tableap1}
  \centering
  \begin{tabular}{cccccc}
    \toprule
    \multirow{2}{*}{Method} & \multicolumn{1}{c}{CIFAR10} & \multicolumn{1}{c}{CIFAR100} 
    &\multirow{2}{*}{Method} & \multicolumn{1}{c}{CIFAR10} & \multicolumn{1}{c}{CIFAR100}\\
    \cmidrule(r){2-3} \cmidrule(r){5-6} 
    & top-1(\%)  & top-1(\%)  &  & top-1(\%)  & top-1(\%) \\
    \midrule
    SEResNet18   &95.02           &78.37          
    &SEResNet18s  &95.34          &78.73           \\
    SEResNet34   &95.56           &78.67           
    &SEResNet34s  &95.74          &80.21           \\
    SEResNet50   &95.27           &79.19           
    &SEResNet50s  &95.76          &80.06           \\
    SEResNet101  &95.31           &79.88           
    &SEResNet101s &95.72          &80.13           \\
    ResNeXt2     &95.67          &79.49            
    &ResNeXt2s    &95.24         &79.55            \\ 
    ResNeXt4     &96.25          &80.23            
    &ResNeXt4s    &96.31         &80.69            \\ 
    Res2NeXt2    &96.01          &81.07             
    &Res2NeXt2s   &96.17         &82.22            \\ 
    Res2NeXt4    &96.12          &79.79             
    &Res2NeXt4s   &96.08         &82.66            \\ 
    \bottomrule
  \end{tabular}
\end{table}

\begin{table}[H]
  \caption{Seed = 1234}
  \label{tableap2}
  \centering
  \begin{tabular}{cccccc}
    \toprule
    \multirow{2}{*}{Method} & \multicolumn{1}{c}{CIFAR10} & \multicolumn{1}{c}{CIFAR100} 
    &\multirow{2}{*}{Method} & \multicolumn{1}{c}{CIFAR10} & \multicolumn{1}{c}{CIFAR100}\\
    \cmidrule(r){2-3} \cmidrule(r){5-6} 
    & top-1(\%)  & top-1(\%)  &  & top-1(\%)  & top-1(\%) \\
    \midrule
    SEResNet18   &95.37           &78.76          
    &SEResNet18s  &95.52          &78.32           \\
    SEResNet34   &95.57           &79.15           
    &SEResNet34s  &95.93          &80.20           \\
    SEResNet50   &95.93           &78.62           
    &SEResNet50s  &95.97          &80.36           \\
    SEResNet101  &95.79           &79.10           
    &SEResNet101s &95.90          &80.06           \\
    ResNeXt2     &95.35          &79.37            
    &ResNeXt2s    &95.52         &79.20            \\ 
    ResNeXt4     &96.97          &81.15            
    &ResNeXt4s    &96.84         &81.06            \\ 
    Res2NeXt2    &95.83          &81.20             
    &Res2NeXt2s   &96.97         &82.01            \\ 
    Res2NeXt4    &96.04          &79.36             
    &Res2NeXt4s   &95.93         &82.71            \\ 
    \bottomrule
  \end{tabular}
\end{table}

\subsection{Params of models}
\label{appm}
\begin{table}[H]
  \caption{FLOPs and model size on CIFAR100}
  \label{tableappa}
  \centering
  \begin{tabular}{cccccccc}
    \toprule
    \multirow{1}{*}{Model} & \multicolumn{1}{c}{FLOPs} & \multicolumn{1}{c}{Params}  
    & \multicolumn{1}{c}{Acc(\%)}
    &\multirow{1}{*}{Model} & \multicolumn{1}{c}{FLOPs} & \multicolumn{1}{c}{Params}  & \multicolumn{1}{c}{Acc(\%)}\\
    \midrule
    ResNet18        &$7.1\times10^{10}$             &11.22M       &76.60   
    &ResNet18s      &$7.2\times10^{10}$             &11.39M       &76.57    \\
    ResNet34        &$1.5\times10^{11}$             &21.33M       &77.09      
    &ResNet34s      &$1.5\times10^{11}$             &21.50M       &77.48      \\
    ResNet50        &$1.7\times10^{11}$             &23.70M       &78.37      
    &ResNet50s      &$1.8\times10^{11}$             &26.48M       &78.65      \\
    ResNet101       &$3.2\times10^{11}$             &42.70M       &77.67     
    &ResNet101s     &$3.4\times10^{11}$             &45.47M       &78.56      \\
    SEResNet18      &$7.1\times10^{10}$             &11.31M       &78.71 
    &SEResNet18s    &$7.2\times10^{10}$             &11.48M       &78.65      \\
    SEResNet34      &$1.5\times10^{11}$             &21.50M       &79.05      
    &SEResNet34s    &$1.5\times10^{11}$             &21.67M       &80.15      \\
    SEResNet50      &$1.7\times10^{11}$             &26.26M       &79.30      
    &SEResNet50s    &$1.8\times10^{11}$             &29.04M       &80.36      \\
    SEResNet101     &$3.2\times10^{11}$             &47.53M       &79.36     
    &SEResNet101s   &$3.4\times10^{11}$             &50.31M       &80.09      \\
    ResNeXt2        &$1.8\times10^{11}$             &9.22M        &79.39    
    &ResNeXt2s      &$1.9\times10^{11}$             &9.89M        &79.49     \\
    ResNeXt4        &$5.4\times10^{11}$             &27.29M       &80.99     
    &ResNeXt4s      &$5.8\times10^{11}$             &29.95M       &81.06      \\
    Res2NeXt2       &$6.1\times10^{11}$             &33.10M       &81.14     
    &Res2NeXt2s     &$6.3\times10^{11}$             &34.61M       &82.22      \\
    Res2NeXt4       &$8.7\times10^{11}$             &47.27M       &79.27     
    &Res2NeXt4s     &$8.9\times10^{11}$             &48.91M       &82.49      \\
    \bottomrule
  \end{tabular}
\end{table}
In table \ref{tableappa}, ResNeXt2 and ResNeXt4 denote $ResNeXt29(2\times64d)$ and $ResNeXt29(4\times 64d)$
respectively. Res2NeXt2 and Res2NeXt4 denote $Res2NeXt29(6c\times24w\times4scales)$ and 
$Res2NeXt29(8c\times25w\times4scales)$ respectively.

It can be found that there is no significant difference in the computational complexity between
the enhanced model and the baseline model.
The accuracy of multiple networks with the new architecture has a stable improvement.

\subsection{additional examples}
\paragraph{Start with recursion formulas}
The recursion formulas of residual networks and the traditional CNNs are
in the following form:
\[X_{i}=F\left(X_{i-1}, \theta_{i}\right)\]
where $X_{i-1}$ and $X_{i}$ denote the input and output of the $i$-th block/layer,
$\theta_{i}$ denotes the parameters of  the $i$-th block/layer.
According to the above equation, 
the output of the current block/layer is only related to the output of the previous block/layer.

\subsubsection{additional example 1}
\begin{figure}[H]
    \centering
    \includegraphics[scale = 1.4]{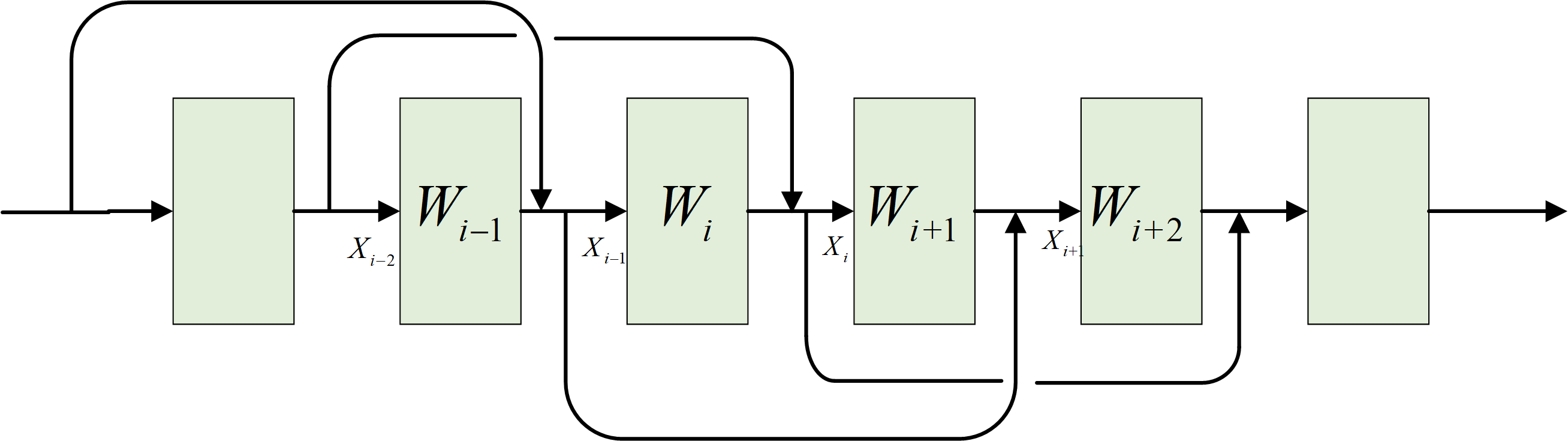}
    \caption{The architecture of additional example 1}
    \label{fig:ads2}
\end{figure}

Consider the following form of recursion formulas:
\[X_{i}=F\left(X_{i-1}, X_{i-2}, \theta_{i}\right)=W_{i} \cdot X_{i-1}+X_{i-2}\]
where $W$ denotes the mapping of the block/layer and $1<i<L+1$.
Based on the recursion formula, a new architecture can be designed as Figure \ref{fig:ads2}.

\subsubsection{additional example 2}
\begin{figure}[H]
    \centering
    \includegraphics[scale = 1.4]{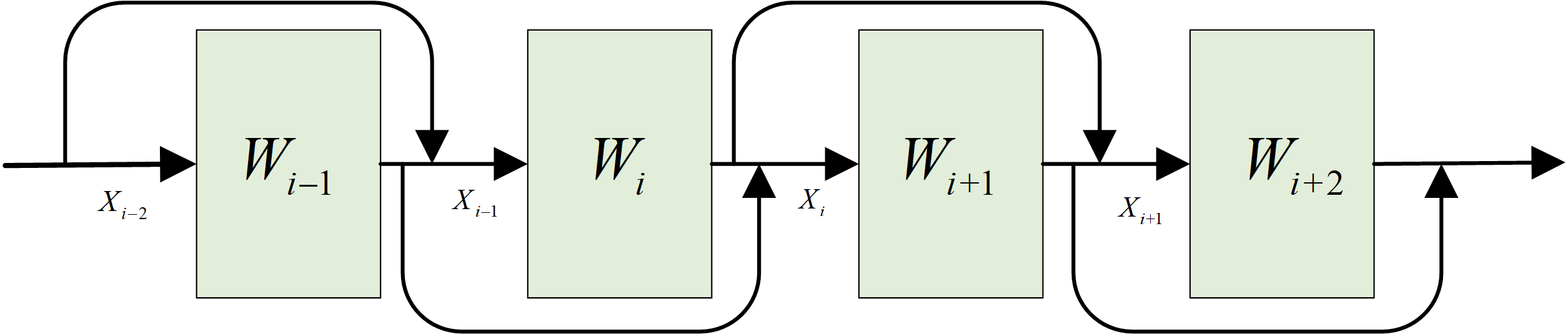}
    \caption{The architecture of additional example 2}
    \label{fig:ads1}
\end{figure}

Consider the following form of recursion formulas:
\[X_{i}=F\left(X_{i-1}, X_{i-2}, \theta_{i}, \theta_{i-1}\right)\]
We can design a recursion formula that fits this form as follows:
\[X_{i}=F\left(X_{i-1}, \theta_{i}\right)+F\left(X_{i-2}, \theta_{i-1}\right)=W_{i} \cdot X_{i-1}+W_{i-1} \cdot X_{i-2}\]
where $W$ denotes the mapping of the block/layer and $1<i<L+1$.
Based on the recursion formula, a new architecture can be designed as Figure \ref{fig:ads1}.

\subsection{Pytorch Code}
\label{ap3}
Here is the Pytorch code of the example in section \ref{example}

\subsubsection{Code of the basic block}
\begin{python}
def forward(self, x):
    out = F.relu(self.bn1(self.conv1(x[1])))
    out = self.bn2(self.conv2(out))
    outp = out
    out += self.shortcut(x[1])
    out = F.relu(out)
    out -= self.shortcut2(x[0])
    out = torch.stack((outp, out))
    return out
\end{python}

\subsubsection{Code of the bottleneck}
\begin{python}
def forward(self, x):
    out = F.relu(self.bn1(self.conv1(x[1])))
    out = F.relu(self.bn2(self.conv2(out)))
    out = self.bn3(self.conv3(out))
    outp = out
    out += self.shortcut(x[1])
    out = F.relu(out)
    out -= self.shortcut2(x[0])
    out = torch.stack((outp, out))
    return out
\end{python}

\subsection{Reproduction details}
\label{ack}

\paragraph{CIFAR} The input size is $32\times32$. All the networks are trained with  
momentum stochastic gradient descent optimizer with a learning rate of 0.1, 
momentum of 0.9, and the weight decay $5\times10^{-4}$. 
Besides, the scheduler used is cosine annealing 
, the number of training epoch is 200 and batch size is 128.

\paragraph{ImageNet} The input size is $224\times224$. All the networks are trained with  
momentum stochastic gradient descent optimizer with a learning rate of 0.1, 
momentum of 0.9, and weight decay $5\times10^{-4}$. 
Besides, the scheduler used is MultiStepLR, the learning rate decays by a factor of 10 every 30 epochs, 
the number of training epoch is 90 and batch size is 256.

\paragraph{Baselines} We do not use any improvement methods, such as label smoothing, etc.
The hyperparameters of the baseline models on the CIFAR and ImageNet are a little different. 

\begin{table}[H]
  \caption{Baseline details on different data sets}
  \label{tablede}
  \centering
  \begin{tabular}{ccccccc}
    \toprule
    \multirow{1}{*}{Dataset} & \multicolumn{1}{c}{Pre} & \multicolumn{1}{c}{Layer1}  
    & \multicolumn{1}{c}{Layer2} &\multicolumn{1}{c}{Layer3} 
    & \multicolumn{1}{c}{Layer4} & \multicolumn{1}{c}{Last} \\
    \midrule
    CIFAR     &3c, $3\times3$, 64c   &64c, 64c       &64c, 128c   
    &128c, 256c      &256c, 512c    &avg-pool(4), fc   \\
    ImageNet   &3c, $7\times7$, 64c   &64c, 64c       &64c, 128c   
    &128c, 256c      &256c, 512c    &avg-pool(7), fc   \\
    \bottomrule
  \end{tabular}
\end{table}

Table \ref{tablede} shows the hyperparameters on different data sets, where
Layer denotes the unit that contains multiple blocks, c denotes channels,
$3\times3$ and $7\times7$ denote convolution kernel size and fc denotes full connection layer.
How many blocks Layer1, Layer2, Layer3 and Layer4 contain determines the number of layers in the network.

And we complete our code with reference to the following open source code:\\
\url{https://github.com/bearpaw/pytorch-classification} with MIT license \\
\url{https://github.com/Res2Net/Res2Net-PretrainedModels}\\
\url{https://github.com/jiweibo/ImageNet} with MIT license\\  
\url{https://github.com/kuangliu/pytorch-cifar} with MIT license

\end{document}